\documentclass{article}
\usepackage{spconf,amsmath,graphicx}

\usepackage{amsthm}
\usepackage{bm}
\usepackage[overload]{empheq}
\usepackage{subcaption}

\usepackage{tikz}

\title{Cell Phantom Video Generation in Elliptical Fourier Descriptor Domain}
\name{
\begin{tabular}{@{}c@{}}
Francesco Benedetto \quad Roberto Basla \quad Luca Magri \quad Giacomo Boracchi
\end{tabular}
}
\address{Department of Electronics Information and Bioengineering, Politecnico di Milano, Italy}

\usepackage{titlesec}
\usepackage{color}

\usepackage[utf8]{inputenc}
\usepackage[english]{babel}
\usepackage[T1]{fontenc} % Font encoding
\usepackage{amssymb} % For \mathbb

\usepackage{graphicx}
\graphicspath{{Images/}} % Path for images' folder
\usepackage{eso-pic} % For the background picture on the title page
\usepackage{caption} % Coloured captions
\usepackage{transparent}

\usepackage{amsmath}
\usepackage{amsthm}
\usepackage{bm}
\usepackage[overload]{empheq}  % For braced-style systems of equations

\usepackage{tabularx}
\usepackage{longtable} % tables that can span several pages
\usepackage{colortbl}
\usepackage{multirow}

\usepackage{algorithm}
\usepackage{algorithmic}

\usepackage[colorlinks=true,linkcolor=black,anchorcolor=black,citecolor=black,filecolor=black,menucolor=black,runcolor=black,urlcolor=black]{hyperref} % Adds clickable links at references
\usepackage{cleveref}
\usepackage{appendix}

\usepackage{enumitem}

\usepackage{amsthm,thmtools,xcolor}
\usepackage{comment}
\usepackage{fancyhdr}
\usepackage{lipsum}
\usepackage{tcolorbox}
\usepackage{stfloats}

\newcommand{\bea}{\begin{eqnarray}}
\newcommand{\eea}{\end{eqnarray}}

\newcommand{\mask}{M}
\newcommand{\maskvideo}{\mathbf{M}}
\newcommand{\synmask}{\widetilde{M}}
\newcommand{\synthmaskvideo}{\widetilde{\mathbf{M}}}

\newcommand{\efd}{z}
\newcommand{\efdTS}{\mathbf{\efd}}
\newcommand{\synthEfdTS}{\widetilde{\efdTS}}

\newcommand{\Dreal}{D_\text{real}}
\newcommand{\Dsynth}{D_\text{synth}}

\newcommand{\encoder}{E\!F\!D}
\newcommand{\decoder}{E\!F\!D^{-1}}

\newcommand{\Generator}{\mathcal{G}_\theta}

\newcommand{\height}{H}
\newcommand{\width}{W}
\newcommand{\maskspace}{\{0,1\}^{\height \times \width}}
\newcommand{\ContourSpace}{\mathbb{R}^{\NCountourPoints \times 2}}
\newcommand{\ContourVideoSpace}{\mathbb{R}^{\NCountourPoints \times 2 \times \MaxTime}}
\newcommand{\EFDSpace}{\mathbb{R}^{4\dnyq \times \MaxTime}}
\newcommand{\VideoSpace}{\{0,1\}^{\height \times \width \times \MaxTime}}

\newcommand{\ContourExtractor}{\mathcal{C}}

\newcommand{\dnyq}{d}              % Nyquist dimension
\newcommand{\ArcLengthParameter}{s}
\newcommand{\XPeriodic}{x(\ArcLengthParameter)}
\newcommand{\YPeriodic}{y(\ArcLengthParameter)}
\newcommand{\HarmonicOrder}{n}

\newcommand{\HarmonicCoeff}[1]{\text{#1}}

\newcommand{\NDiffusionSteps}{K}
\newcommand{\DiffusionStep}{k}

\newcommand{\LossFunction}{\mathcal{L}}
\newcommand{\PredictedSignal}{\hat{\efdTS}}

\newcommand{\TimeStep}{t}
\newcommand{\MaxTime}{T}

\newcommand{\RealSamplesIndex}{i}
\newcommand{\RealSamplesSize}{S_{\text{real}}}
\newcommand{\SynthSamplesIndex}{j}
\newcommand{\SynthSamplesSize}{S_{\text{synth}}}

\newcommand{\NCountourPoints}{N}

\newcommand{\ellipse}{e}

\newcommand{\Contour}{C}
\newcommand{\ContourVideo}{\mathbf{\Contour}}
\newcommand{\SynthContourVideo}{\mathbf{\widetilde{\Contour}}}

\newcommand{\solidity}{\mathtt{s}}
\newcommand{\overlapRatio}{\mathtt{o}}

\newcommand{\hOneSpacePre}{\vspace{-2mm}}
\newcommand{\hOneSpacePost}{\vspace{-2mm}}
\newcommand{\hTwoSpacePre}{\vspace{-2mm}}
\newcommand{\hTwoSpacePost}{\vspace{-1mm}}

\newcommand{\eqSpacePre}{\vspace{-2mm}}
\newcommand{\eqSpacePost}{

\vspace{-2mm} \noindent}
\newcommand{\eqShortSpacePost}{

\vspace{-1mm} \noindent}

\ninept

\begin{document}
\maketitle

\begin{tikzpicture}[remember picture,overlay]
  \node[anchor=south, yshift=20pt] at (current page.south) {
    \parbox{\dimexpr\textwidth-\fboxsep-\fboxrule\relax}{
      \centering \scriptsize 
      \copyright~2026 IEEE. Personal use of this material is permitted.  Permission from IEEE must be obtained for all other uses, in any current or future media, including reprinting/republishing this material for advertising or promotional purposes, creating new collective works, for resale or redistribution to servers or lists, or reuse of any copyrighted component of this work in other works.
    }
  };
\end{tikzpicture}

\begin{abstract}
Training Deep Neural Networks for tracking individual cells in biomedical videos requires a large amount of annotated data. The annotation of videos for cell tracking is very time consuming and often requires domain expertise; this explains the limited availability of public annotated data to address important medical problems like tissue repair or cancer treatment. Generating synthetic videos along with their Ground Truth annotations is a promising solution that relies, as a foundational first step, on the synthesis of single cell annotations (or \textit{phantoms}). Phantoms need to be time consistent, as they have to replicate biological processes that are specific to the cell types. In this work, we propose a novel framework for generating videos of cell phantoms in the Elliptical Fourier Descriptors (EFDs) domain, a compact and geometrically interpretable representation for 2D closed contours. We represent the cell phantom evolution as a multivariate time series of EFD coefficients, introducing a strong prior for cell morphology and enabling the efficient generation of sequences that evolve coherently in time. Our experimental validation proves that modelling the temporal evolution in EFD space enables the generation of biologically plausible phantom videos. Our method can be used in generative pipelines for synthesizing annotated data for cell tracking, thus strongly mitigating the annotation effort for creating new datasets. Our code is available for download here:
https://github.com/FrancescoBenedetto99/efd-cell-video-gen.
\end{abstract}

\begin{keywords}
Cell Phantom, Cell Tracking, Biomedical Video Generation
\end{keywords}

\hOneSpacePre
\section{Introduction}
\begin{figure}[t]
\centering
\includegraphics[width=.8\columnwidth]{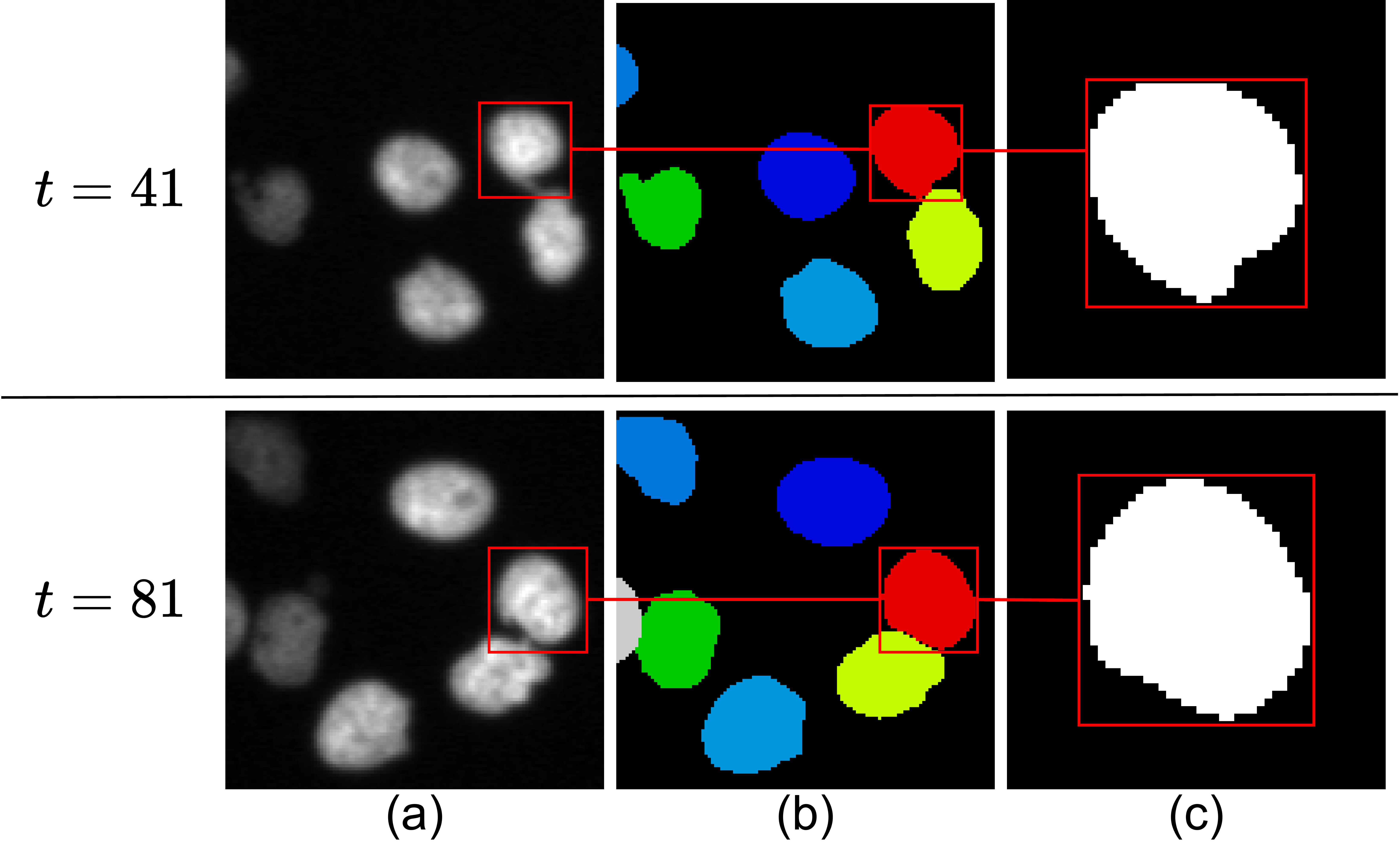}
\vspace{-3mm}
\caption{Frames from the Fluo-N2DL-HeLa dataset \cite{neumann_phenotypic_2010} at different time steps $\TimeStep$. (a) Histological video, (b) cell tracking Ground Truth, and (c) a cell phantom. The bounding box highlights the same cell at different times $\TimeStep = 41$, $t = 81$.}
\label{fig:Microscopy and segmentation}
  \vspace{-6mm}
\end{figure}

\hOneSpacePost
Cell tracking is the problem of analysing videos of living cells (Fig. \ref{fig:Microscopy and segmentation}a) and segment and track each cell instance along the frames \cite{chakraborty_conditional_2014} (Fig. \ref{fig:Microscopy and segmentation}b). State-of-the-art solutions in cell tracking are represented by Deep Learning (DL) models \cite{maska_cell_2023}, which are becoming fundamental tools to study the properties of living cells and to understand tissue development and repair, cancer treatment and drug development \cite{yazdi_survey_2024}. Unfortunately, training DL models for cell tracking requires large datasets annotated by domain experts, which are very time consuming to prepare. Moreover, the variability in cell types (or \textit{cell lines}) and tissue preparation procedures for microscopy imaging may require annotating videos for each setting. An appealing perspective to train cell tracking models consists in synthetically generating custom annotated datasets, which is nowadays possible given the advances in data generation. In this work, we address the problem of generating Ground Truth (GT) videos of binary masks for single cells (also called \textit{cell phantoms}, Fig. \ref{fig:Microscopy and segmentation}c) in cell tracking settings. 

Research on the generation of synthetic annotated datasets for cell segmentation and tracking follows two main paradigms: \textit{end-to-end} and \textit{pipeline-based}. End-to-end approaches use Deep generative models (e.g., GANs or diffusion models) to directly synthesize complete microscopy images and videos with their annotations in a single step \cite{dimitrakopoulos_ising-gan_2020}. Although achieving high visual realism, these data-driven, black-box methods require large annotated datasets themselves and offer limited control over individual cell properties (morphology, positioning, movements and texture), losing control on the generation process. On the other hand, pipeline-based approaches explicitly decompose generation into multiple stages, allowing the injection of domain expertise in the generation process. Pipelines generating images are typically made of three steps: (1) phantom generation, (2) spatial positioning, and (3) texture image synthesis, enabling better control over each generation aspect~\cite{basla_expert-driven_2024}. In videos, a straightforward extension of these pipelines for the accounting of cell evolution and movement is (1) \textit{phantom videos generation}, (2) \textit{composition, movement and interaction modelling}, and (3) \textit{video texture synthesis}. This paper proposes a method to address the essential first step.

So far, both end-to-end and pipeline-based approaches focused primarily on the generation of realistic textures for images and videos, often paying less attention to phantom generation. However, phantoms determine the quality of video-level annotations and consequently of the generated dataset. Phantoms must capture the macroscopic properties of the cell, namely the shape and size, and need to be morphologically realistic. Thus, phantoms should consist of a single, closed-contour and \textit{simply connected} component (i.e., without disjoint parts nor holes), and should evolve coherently over time. The mainstream approaches  \cite{bahr_cellcyclegan_2021, eschweiler_3d_2021, fernandez_annotated_2025} use phantoms generated by a Statistical Shape Model (SSM). In particular, CellCycleGAN \cite{bahr_cellcyclegan_2021} models a Markov process over the cell life cycle stages (that go from birth to division, or mitosis) to define more precisely the morphology, but it requires phantoms to be additionally annotated with their cell state. Another limitation of \cite{bahr_cellcyclegan_2021} is the description of the coordinates of the phantom's contours with a Principal Component Analysis (PCA), often resulting in a completely data-driven process that is constrained by the dataset and lacks an interpretable geometric structure. Our work addresses the limitations of SSM-based phantom generation by proposing a generative framework that learns to synthesize videos of individual cell phantoms according to the cellular dynamics.  Our method generates realistic phantom videos, enabling the following stages of a cell tracking dataset generation pipeline to model cell positioning and texture synthesis.
\begin{figure*}[t]
  \centering
  \includegraphics[width=.95\textwidth]{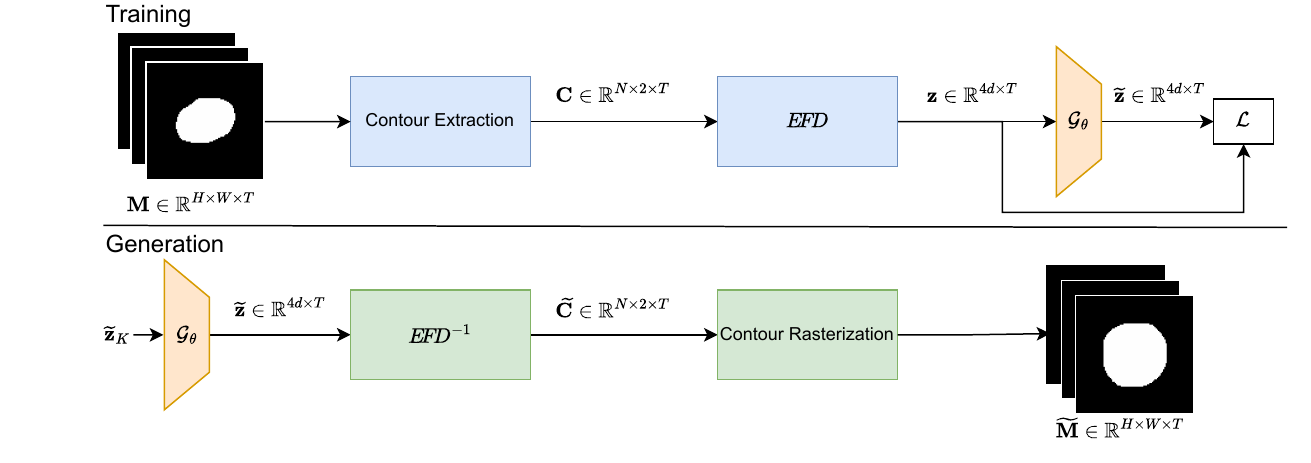}
  \vspace{-5mm}
  \caption{Our pipeline for phantom video generation. (Top) During training we extract a contour video $\ContourVideo$ from each binary phantom video $\maskvideo$, then encode $\ContourVideo$ in the EFD domain to obtain the time series $\efdTS$. We train a generative model $\Generator$ minimizing the loss $\LossFunction$ between real EFD time series $\efdTS$ and generated time series $\synthEfdTS$ to learn the temporal evolution of cell phantoms in EFD space. (Bottom) During generation, we unconditionally sample synthetic EFD time series $\synthEfdTS$ from the trained model starting from a noise vector $\synthEfdTS_\NDiffusionSteps$. We decode $\synthEfdTS$ into a synthetic contour $\SynthContourVideo$ which, once rasterized, corresponds to a synthetic mask video $\synthmaskvideo$.}
  \label{fig:pipeline}
    \vspace{-3mm}
\end{figure*}

In particular, we generate phantom videos in the domain of Elliptical Fourier Descriptors (EFDs), which provide a compact representation of cell phantoms. We encode each video frame in few coefficients representing ellipses that approximate the cell's contour. Therefore, we obtain a collection of multivariate time series in the compact EFD space describing the cellular evolution that we use for training Diffusion-TS \cite{yuan_diffusion-ts_2024}, a Diffusion Model specific for multivariate time series. Diffusion-TS explicitly disentangles the trend and seasonality components of a time series, learning the long-term deformations of each cell phantom and its higher-frequency shape variations.
% We thus leverage the compactness of EFD representation and the realism of state-of-the-art generative models in the learning process. 
We then unconditionally synthesize EFD time series that, once reconstructed through the inverse EFD transform, correspond to realistic-looking cell phantom evolutions in pixel space. Indeed, the EFD representation has great advantages for cell phantom modelling: it naturally enforces geometrical constraints such as the presence of only one simply connected component, allows the representation of phantoms using few parameters, and enables an effective generative model to operate on simpler signals. To the best of our knowledge, our work is the first to apply EFDs in video generation settings, with prior research limited to static phantom generation in 2D or 3D \cite{scalbert_generic_2019, al-thelaya_inshade_2021}. 

In our experiments, we verify that our approach enjoys two fundamental properties: morphological statistical consistency and temporal coherence. We show that phantom videos generated in EFD space closely follow the evolution of morphological descriptors of real data in time. We compare our phantoms against the SSM generative method in CellCycleGAN, highlighting 
% a higher quality from morphological and qualitative 
more realistic morphological properties in both a quantitative evaluation and a qualitative comparison over a dataset used in both works. Additionally, we generate phantom videos starting from the Fluo-N2DL-HeLa dataset of the Cell Tracking Challenge \cite{maska_cell_2023} (CTC Dataset), assessing the generalization power of our method over differently-evolving cell lines.
The morphological comparison with real data suggests that our phantom generation method can be integrated in a pipeline-based framework for generating diverse cell tracking datasets.

\hOneSpacePre
\section{Problem Formulation}
\hOneSpacePost

\label{sec:problem_formulation}
We focus on phantom video generation, i.e., the generation of the binary mask of a single, simply connected component representing the cell's morphology in time. We assume that a real dataset $\Dreal$ of cell phantom videos $\maskvideo$ is provided:
\eqSpacePre
\begin{equation}
    \Dreal = \{\maskvideo^{(\RealSamplesIndex)}\}_{\RealSamplesIndex=1}^{\RealSamplesSize},
\end{equation}
\eqSpacePost where each video $\maskvideo^{(\RealSamplesIndex)}$ is a sequence of 2D binary images containing a single phantom each:
\eqSpacePre
\begin{equation}
    \maskvideo^{(\RealSamplesIndex)} = (\mask^{(\RealSamplesIndex)}_\TimeStep)_{\TimeStep=1}^{\MaxTime_\RealSamplesIndex}, \quad \mask^{(\RealSamplesIndex)}_\TimeStep \in \maskspace,
\end{equation}
\eqSpacePost where $\height$ and $\width$ are the spatial dimensions, and $\maskvideo^{(\RealSamplesIndex)}$ is a tuple of binary masks with ${\MaxTime_\RealSamplesIndex}$ frames. We assume a limited number of $\RealSamplesSize$ samples is available as in realistic biomedical scenarios. Our objective is to generate a synthetic dataset:
\eqSpacePre
\begin{equation}
\Dsynth = \{ \synthmaskvideo^{(\SynthSamplesIndex)} \}_{\SynthSamplesIndex=1}^{\SynthSamplesSize}, \quad \synthmaskvideo^{(\SynthSamplesIndex)} = (\synmask^{(\SynthSamplesIndex)}_{\TimeStep})_{\TimeStep=1}^{{\MaxTime_\SynthSamplesIndex}},
\end{equation}
\eqSpacePost with $\synmask^{(\SynthSamplesIndex)}_{\TimeStep} \in \maskspace$, such that $\SynthSamplesSize \gg \RealSamplesSize$. Additionally, phantom variations across frames should be coherent with the morphology of the specific cell line in $\Dreal$ and the temporal interval between two consecutive frames. For ease of notation, in the following we omit the dataset indices ($\RealSamplesIndex$ and $\SynthSamplesIndex$).

\hOneSpacePre
\section{Method}
\hOneSpacePost
\label{sec:method}
As illustrated in Fig. \ref{fig:pipeline}, our method is organized in two main phases: model training and synthetic dataset generation. For training, we first extract the contour from each frame of a phantom video and encode it in EFD space, using this representation to train a generative model that learns the phantom evolution in time. During generation, we unconditionally generate new EFD time series to be converted first into contours and, finally, in phantom videos.

\hTwoSpacePre
\subsection{EFD Encoding of Phantom Videos}
\begin{figure}[t]
\centering
\includegraphics[width=.9\columnwidth]{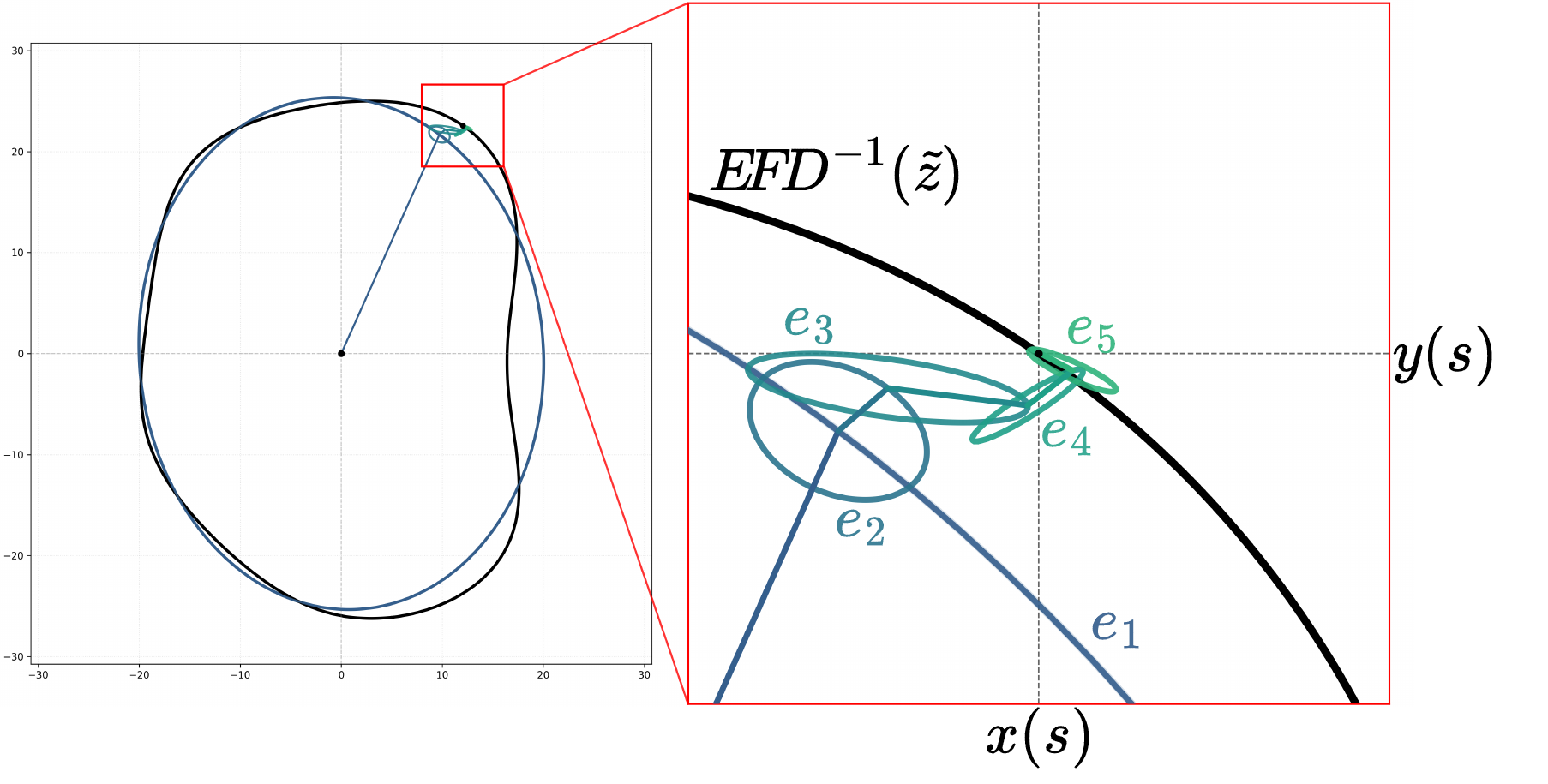}
\vspace{-4mm}
\caption{The $\encoder$ transform parametrizes the points of each phantom's contour as a function of $\ArcLengthParameter \in [0, 2\pi]$, representing it through the coordinates $\XPeriodic$ and $\YPeriodic$. Then, it traces a chain of ellipses ($\ellipse_1 \dots \ellipse_5$) that add increasing detail to the reconstructed contour (in black). The contour is centered in $(0, 0)$ for invariance with respect to the position.}
\label{fig:epicycle}
\end{figure}

\hTwoSpacePost
\label{sec:efd_encoding}
We map each mask video $\maskvideo$ containing a single cell phantom in a multivariate time series $\efdTS \in \EFDSpace$ where each frame is encoded using $\dnyq$ ellipses that approximate the cell contour (Fig. \ref{fig:epicycle}). To this purpose, we first apply a contour extraction operator $\ContourExtractor$ that, for each frame at time steps $\TimeStep\in[1,\dots,\MaxTime]$, \textit{(i)} extracts the contour pixels that represent a closed curve, and \textit{(ii)} applies geodesic uniform resampling via arc-length parametrization for obtaining a better spectral representation \cite{al-thelaya_inshade_2021}. This process maps each binary frame $\mask_\TimeStep$ to the set of $(x, y)$ coordinates of the $\NCountourPoints$ contour points of the shape:
\eqSpacePre
\begin{equation}
    \ContourExtractor: \maskspace \rightarrow \ContourSpace.
\end{equation}
\eqSpacePost Thus, we obtain the contour video
\eqSpacePre
\begin{equation}
    \ContourVideo = (\ContourExtractor(\mask_{\TimeStep}))_{\TimeStep=1}^{\MaxTime}, \qquad \forall \mask_{\TimeStep} \in \maskvideo,
\end{equation}
\eqSpacePost where $\ContourVideo \in \ContourVideoSpace$. We take $\NCountourPoints = 128$ to select enough details for contour sampling.

Cell contours $\ContourVideo$ are closed loops with the points' coordinates that are continuous, periodic functions $\XPeriodic$ and $\YPeriodic$ of the normalized arc-length parameter $\ArcLengthParameter \in [0, 2\pi]$. This periodicity allows us to decompose the shape using the Elliptical Fourier Transform $\encoder$ \cite{kuhl_elliptic_1982}. For each harmonic order $\HarmonicOrder$, the specific frequency component of $\XPeriodic$ (weighted by coefficients $\HarmonicCoeff{a}_\HarmonicOrder, \HarmonicCoeff{b}_\HarmonicOrder$) and $\YPeriodic$ (weighted by $\HarmonicCoeff{c}_\HarmonicOrder, \HarmonicCoeff{d}_\HarmonicOrder$) are combined. While these components represent simple 1D oscillations individually, their superposition in the 2D plane traces a unique ellipse.  The full contour is approximated by summing the centroid $(A_0, C_0)$ and all the harmonic ellipses up to a truncation order $\dnyq$ which represents the number of selected ellipses:
\eqSpacePre
\begin{equation}
\label{eq:efd_reconstruction}
    \begin{bmatrix} \XPeriodic \\ \YPeriodic \end{bmatrix} = 
    \underbrace{\begin{bmatrix} A_0 \\ C_0 \end{bmatrix}}_{\text{Centroid}} + 
    \sum_{n=1}^{\dnyq} 
    \underbrace{
    \begin{bmatrix} 
    \HarmonicCoeff{a}_\HarmonicOrder \cos(\HarmonicOrder \ArcLengthParameter) + \HarmonicCoeff{b}_\HarmonicOrder \sin(\HarmonicOrder \ArcLengthParameter) \\ 
    \HarmonicCoeff{c}_\HarmonicOrder \cos(\HarmonicOrder \ArcLengthParameter) + \HarmonicCoeff{d}_\HarmonicOrder \sin(\HarmonicOrder \ArcLengthParameter) 
    \end{bmatrix}
    }_{\text{$\HarmonicOrder$-th harmonic ellipse}}.
\end{equation}
\eqSpacePost Fig. \ref{fig:epicycle} shows this EFD decomposition: the sum in \eqref{eq:efd_reconstruction} corresponds to a chain of epicycles where the center of the harmonic ellipse $\ellipse_\HarmonicOrder$ of order $\HarmonicOrder$ travels along the perimeter of the ellipse $\ellipse_{\HarmonicOrder - 1}$ (see red box). The first harmonic captures the overall shape and orientation (best-fitting ellipse $e_1$), while higher-order harmonics progressively add finer morphological details. In our encoding, we translate the center of $\ellipse_1$ to the origin, namely we set $A_0=C_0=0$ for invariance to the cell position. Further details on EFD transforms can be found in \cite{kuhl_elliptic_1982}.

We encode the phantom contour at frame $\TimeStep$ as a spectral vector $\efd_\TimeStep \in \mathbb{R}^{4\dnyq}$ by concatenating the EFD coefficients:
\eqSpacePre
\begin{equation}
    \efd_\TimeStep = [\HarmonicCoeff{a}_1, \HarmonicCoeff{b}_1, \HarmonicCoeff{c}_1, \HarmonicCoeff{d}_1, \dots, \HarmonicCoeff{a}_\dnyq, \HarmonicCoeff{b}_\dnyq, \HarmonicCoeff{c}_\dnyq, \HarmonicCoeff{d}_\dnyq]^\top.
\end{equation}
\eqShortSpacePost The entire video evolution is thus mapped to the multivariate time series $\efdTS = [\efd_1, \dots, \efd_\MaxTime] \in \EFDSpace$. We normalize values in $\efdTS$ between 0 and 1 for easier training of the generative model. We also determine the number of ellipses $\dnyq$ via Power Spectral Density (PSD) analysis on the training set, selecting the minimum $\dnyq$ that retains 99.99\% of the cumulative power, yielding $\dnyq = 9$. 

\begin{figure}[t]
\centering
\includegraphics[width=.75\columnwidth]{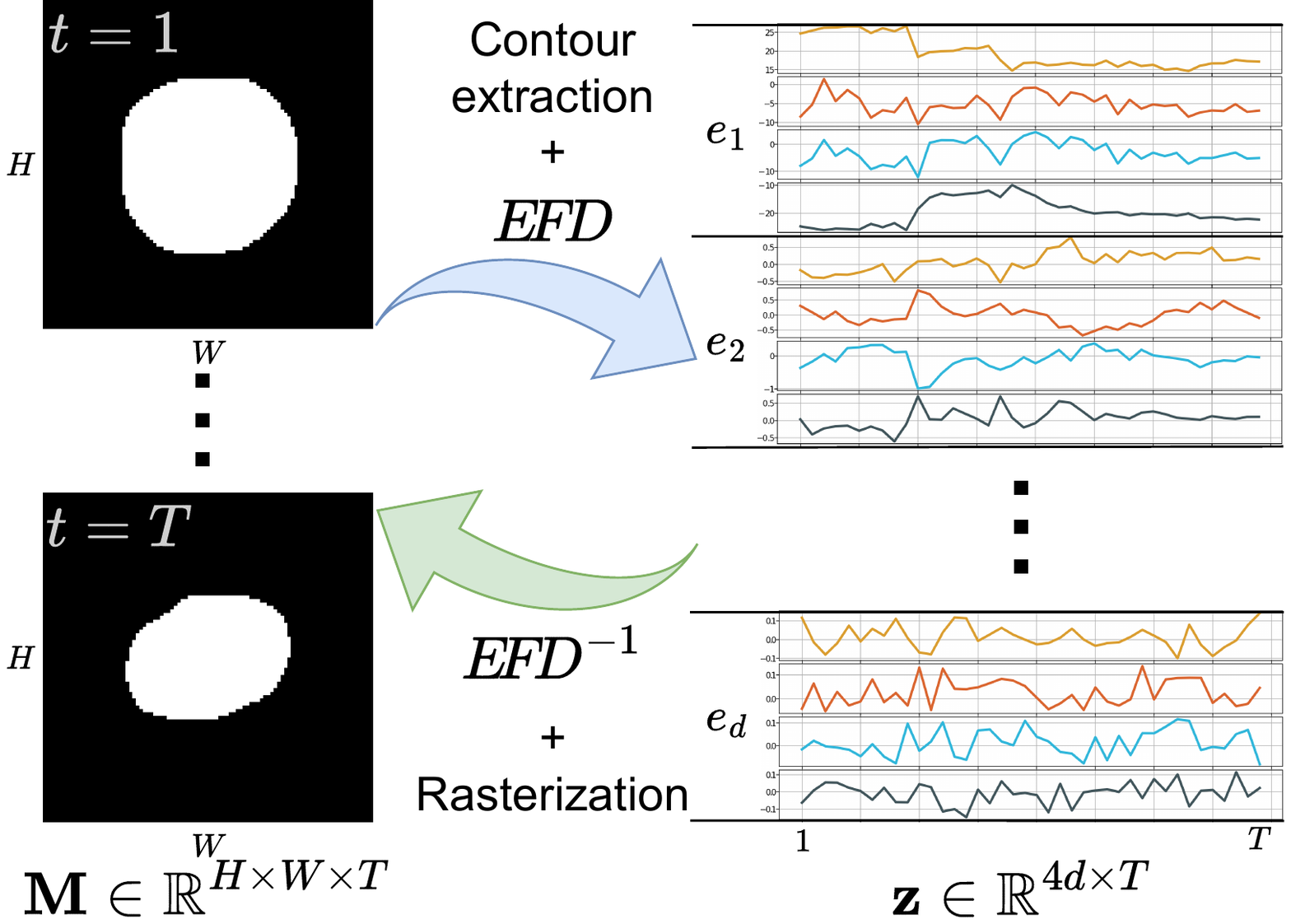}
\vspace{-5mm}
\caption{We encode each mask video $\maskvideo$ into a time series representation $\efdTS$ by first extracting each frame's contour video $\ContourVideo$ and then encoding it with the $\encoder$ transformation. The multivariate time series $\efdTS$ encodes $\dnyq$ ellipses ($\ellipse_1 \dots \ellipse_\dnyq$), each  represented by four coefficients for each time step $\TimeStep$. The inverse transformation $\decoder$, coupled with the contour rasterization, reconstructs the video.}
\label{fig:efd_transform}
\vspace{-4mm}
\end{figure}

This EFD transform offers two key advantages. First, it reduces the dimensionality of each frame by mapping the $\height \times \width$ pixel grid into a compact descriptor space of size $4\dnyq$ for a total of $4 \cdot 9 = 36$ parameters while a $128 \times 128$ mask corresponds to $16384$ pixels. Second, it enforces a topological prior: unlike raw segmentation masks which may contain holes or disconnected artifacts due to noise, the truncated Fourier reconstruction is guaranteed to be smooth and yield a closed boundary with a simply connected phantom. The encoding process is illustrated in Fig. \ref{fig:efd_transform} (top), where the contour extraction and $\encoder$ decomposition yield a multivariate time series $\efdTS \in \EFDSpace$ corresponding to the phantom video. 

\hTwoSpacePre
\subsection{Modelling EFD Multivariate Time Series}
\hTwoSpacePost
\label{sec:diffusion_model}
To generate realistic phantom evolutions, we model the probability distribution of time series in EFD space. We employ the Diffusion-TS framework \cite{yuan_diffusion-ts_2024}, a state-of-the-art Diffusion Model designed for time series that is particularly effective as it decouples the trend (or long-term variation) and seasonality (or periodic fluctuations) components. As per the Diffusion Model paradigm, the training involves a forward diffusion process that gradually adds Gaussian noise to a real EFD multivariate time series $\efdTS_0 \in \EFDSpace$ over $\NDiffusionSteps$ diffusion steps. At step $\DiffusionStep$, the noisy signal $\efdTS_\DiffusionStep$ is defined as:
\eqSpacePre
\begin{equation}
    \efdTS_\DiffusionStep = \sqrt{\bar{\alpha}_\DiffusionStep}\efdTS_0 + \sqrt{1-\bar{\alpha}_\DiffusionStep}\epsilon, \quad \epsilon \sim \mathcal{N}(0, \mathbf{I})
\end{equation}
\eqShortSpacePost
where $\bar{\alpha}_\DiffusionStep$ follows a cosine noise schedule to preserve signal structure until later steps, and $\efdTS_\DiffusionStep$ has the same dimensionality as $\efdTS_0$.

The generative model $\Generator$ learns to reverse the diffusion process through a Transformer encoder-decoder architecture but, unlike Diffusion Models for images that at iteration $\DiffusionStep$ predict the noise $\epsilon$ to be removed, Diffusion-TS directly predicts the clean signal $\PredictedSignal_0$ from the noisy input $\synthEfdTS_\DiffusionStep$. For the next diffusion step, the model computes the signal $\synthEfdTS_{\DiffusionStep-1}$  as a weighted linear combination of $\PredictedSignal_0$, $\synthEfdTS_\DiffusionStep$ and a Gaussian noise term for stochasticity.  Moreover, the model ensures high fidelity in the spectral domain by optimizing a hybrid loss function $\mathcal{L}$. This loss promotes the disentangling of temporal dynamics by combining a Mean Squared Error in the time domain with a loss term based on the Fast Fourier Transform $\mathcal{F}$ applied to the signal as per the original work \cite{yuan_diffusion-ts_2024}:
\eqSpacePre
\begin{equation}
    \mathcal{L} = \mathbb{E}_{\DiffusionStep, \efdTS_0} \left[ \lambda_1 ||\efdTS_0 - \PredictedSignal_0||^2 + \lambda_2 ||\mathcal{F}(\efdTS_0) - \mathcal{F}(\PredictedSignal_0)||^2 \right],
\end{equation}
\eqShortSpacePost
where $\PredictedSignal_0 = \Generator(\efdTS_k, k)$ is the reconstructed time series.

\hTwoSpacePre
\subsection{Synthetic Video Generation}
\hTwoSpacePost
During inference (Fig. \ref{fig:pipeline}, bottom), generation starts by sampling a noise seed $\synthEfdTS_\NDiffusionSteps \in \EFDSpace$ representing a sequence of pure Gaussian noise $\synthEfdTS_\NDiffusionSteps \sim \mathcal{N}(0, \mathbf{I})$. Then, $\Generator$ iteratively denoises $\synthEfdTS_\NDiffusionSteps$ for $\NDiffusionSteps = 200$ steps, obtaining a synthetic EFD time series $\synthEfdTS$. We then apply the inverse transformation $\decoder$ to map the generated time series $\synthEfdTS$ into the contour coordinates $\SynthContourVideo \in \ContourVideoSpace$, which are finally rasterized into binary mask videos $\synthmaskvideo \in \VideoSpace$ (Fig. \ref{fig:efd_transform}, bottom).

\hOneSpacePre
\section{Experiments}
\hOneSpacePost
\label{sec:experiments}
We assess the realism of our generated phantoms by comparing their morphological properties against real phantom videos.  We use the Compyda tool \cite{necasova_compyda_2024} to quantitatively compare real and synthetic datasets.

\begin{table}[t]  
\centering
\caption{Evaluation datasets information. We evaluate the properties of the generated phantoms against the test set.}
\label{tab:datasets}
\scriptsize
\begin{tabular}{l|c|c|c|c|c}

\textbf{Dataset}     & $\boldsymbol{\height \times \width}$ & $\boldsymbol{\MaxTime}$ & $\boldsymbol{\RealSamplesSize}$ & \textbf{\#Train} & \textbf{\#Test} \\
\hline
 CellCycleGAN dataset & $96 \times 96$& 40& 326& 258&68\\
 CTC dataset & $128 \times 128$& 50& 2954& 2652 & 302\\
\end{tabular}
\vspace{-4mm}
\end{table}

\hTwoSpacePre
\subsection{Datasets and Preprocessing}
\hTwoSpacePost
\label{sec:datasets}
In order to evaluate the generation power of our method, we generate phantoms starting from two different datasets described in Table \ref{tab:datasets}. 

\begin{figure*}[t]
    \centering
    \includegraphics[width=0.95\textwidth]{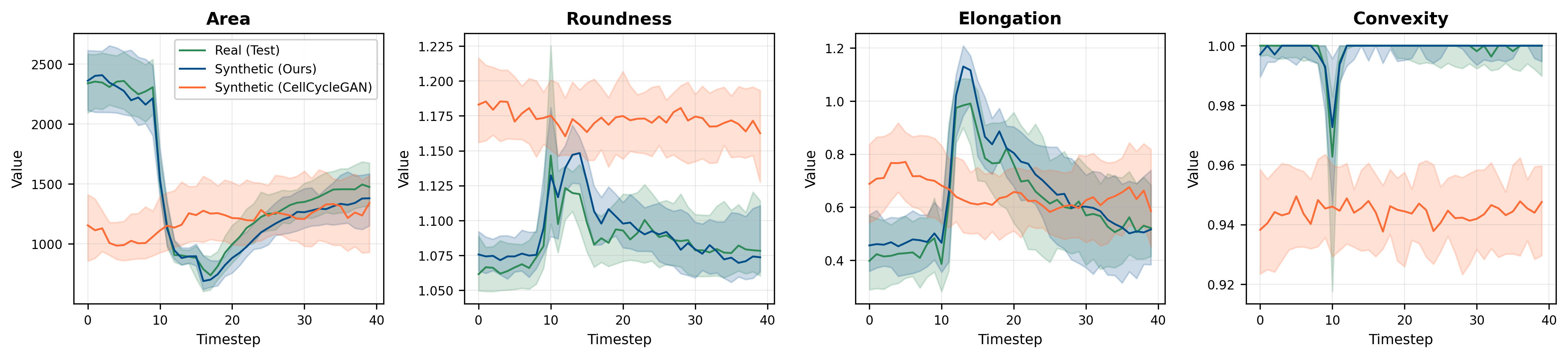}
    \vspace{-3mm}
    \caption{Temporal evolution of selected morphological features on the CellCycleGAN dataset. Solid lines represent population means over synchronized phantom videos, while shaded regions show interquartile ranges. We report the real test set evolution (green), the phantoms generated by our method (blue) and generated by the SSM (orange). The sudden drop in the area metric for the test set corresponds to mitotic events.
    %as, for the comparison dataset, a mitosis is treated by continuing tracking one daughter cell.
    }
    \label{fig:temporal_features}
    \vspace{-4mm}
\end{figure*}

\begin{figure}[t]
\centering
\includegraphics[width=.8\columnwidth]{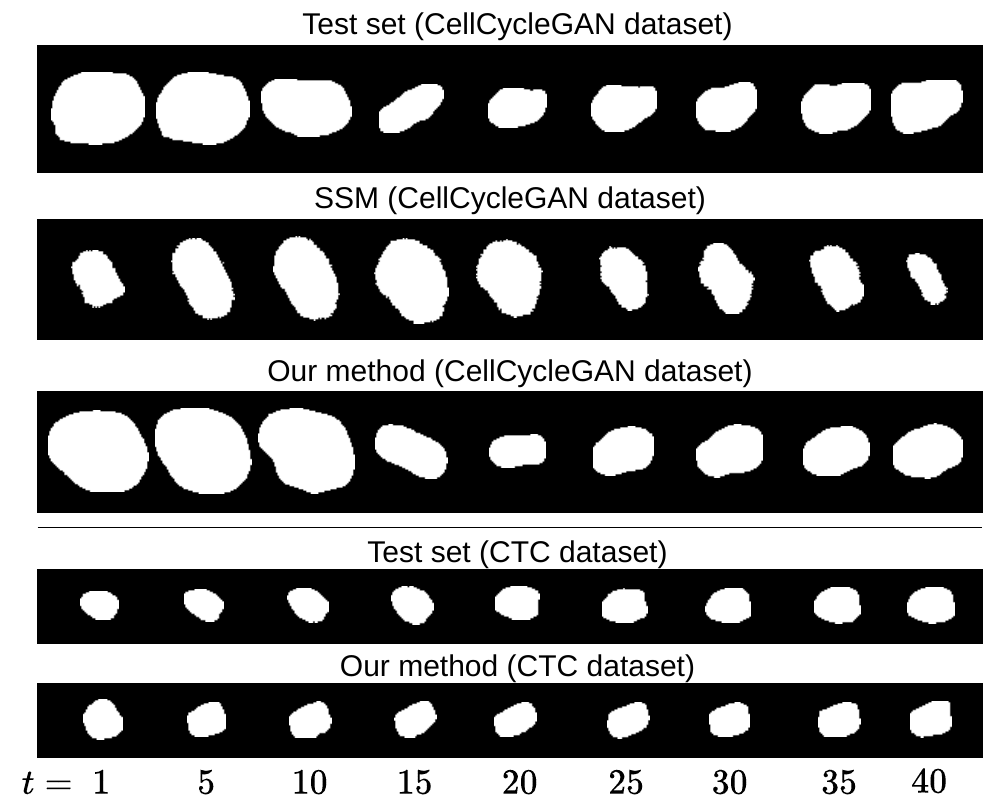}
\vspace{-2mm}
\caption{Real and synthetic cell phantom videos at intervals of $5$ frames. In the CellCycleGAN dataset, mitoses happen after $\TimeStep = 10$. Our method replicates the cell cycle dynamics as modelled in the training set and generates more regular and realistic phantoms than CellCycleGAN's SSM \cite{bahr_cellcyclegan_2021}. We can also replicate the phantom evolution in the CTC dataset. In both cases, thanks to the $\encoder$ encoding phase, our contours are smooth.}
\label{fig:qualitative}
\vspace{-4mm}
\end{figure}

\vspace{-3mm}
\paragraph{CellCycleGAN dataset} This dataset \cite{zhong_unsupervised_2012} contains annotated videos of cell evolutions with both the phantom and the cell cycle stage. All phantom videos in the dataset are synchronized over a common cell state. However, during and after mitotic events, the annotation follows only the daughter cell closest to the parent's centroid, discarding the other. Consequently, each video $\maskvideo$ represents a continuous evolution of a cell phantom which remains a single simply connected component throughout the $\MaxTime = 40$ frames. In Fig. \ref{fig:temporal_features} (Real) and \ref{fig:qualitative} (Test set), videos synchronized over cell states exhibit an area drop due to mitotic events. Since the original data consists of centered cell phantom videos, these data perfectly align with our problem formulation, allowing modelling the entire cell cycle without explicitly handling the division of cells.

We compare our solution against CellCycleGAN \cite{bahr_cellcyclegan_2021}, which is trained on this dataset, enabling a fair comparison. Since the dataset groups cell phantom videos in 7 GT histological sequences, we use 5 for training and 2 for testing  
% We use phantoms from 5 histological GT sequences for training and those from the remaining 2 for testing 
corresponding to a 80/20 train-test split.

\vspace{-3mm}
\paragraph{CTC dataset} The Fluo-N2DL-HeLa dataset \cite{neumann_phenotypic_2010} from the Cell Tracking Challenge contains annotated biomedical videos (see Fig. \ref{fig:Microscopy and segmentation}a and \ref{fig:Microscopy and segmentation}b) with 92 frames. We extract cell phantoms from the given \emph{Silver Truth} provided in the CTC and filter out incomplete instances like cells touching the image border by more than 10 pixels (deemed partially out of frame) or having a high overlap with other cells. This processing removes incomplete cells which we identify by computing the solidity $\solidity$ and the overlap ratio $\overlapRatio_{\RealSamplesIndex,\RealSamplesIndex'}$ between two instances $\RealSamplesIndex$ and $\RealSamplesIndex'$ as:
\eqSpacePre
\begin{equation}
    \solidity_{\RealSamplesIndex} =\frac{|\mask_\RealSamplesIndex|}{|\text{ConvHull}(\mask_{\RealSamplesIndex})|},
    \quad
    \overlapRatio_{\RealSamplesIndex,\RealSamplesIndex'} = \frac{|\text{ConvHull}(\mask_i) \cap \mask_{\RealSamplesIndex'}|}{|\text{ConvHull}(\mask_i)|}.
\end{equation}
\eqShortSpacePost
We discard cells for having high overlap if $\solidity_\RealSamplesIndex < 0.969$ and $\overlapRatio_{\RealSamplesIndex,\RealSamplesIndex'} > 0.017$. We manually set these thresholds to retain well-segmented cells while avoiding excessive data reduction. Since this outlier removal process may interrupt the cell's temporal evolution, we extract uninterrupted blocks of 50 frames for each phantom using a sliding window approach. From the original 924 phantom videos of varying lengths, this extraction yields 2954 videos with $\MaxTime = 50$ (see Table \ref{tab:datasets}). Differently from the CellCycleGAN dataset, phantoms do not continue after mitosis and cell states are not synchronized. Therefore, we model the mitotic cycle only up to the cell division. 

The CTC dataset also includes lineage information, i.e., the tracking GT that includes relationships between parent and daughter cells in the video. To prevent data leakage, we account for lineages in the train-test split assigning entire lineages to one split.
%, since cells from the same lineage could share morphological information. 
This process guarantees that no cell track appears in both train and test since all descendants of a cell belong to the same split as their ancestors.

\begin{figure}[t]
    \centering
    \includegraphics[width=\columnwidth]{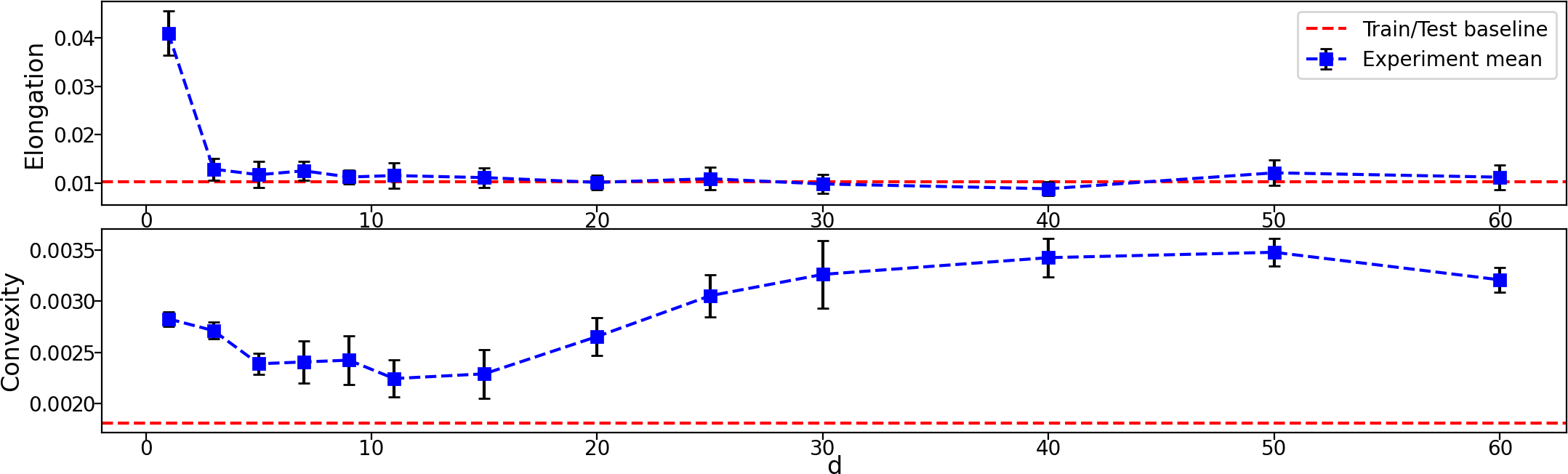}
    \vspace{-7mm}
    \caption{Example of performance variation for selected metrics and values of $\dnyq \in [1, 60]$. Providing enough harmonics, our method reaches a plateau of performance close to the real data baseline in the elongation case (top), while convexity reports a minimum close to the value of $\dnyq = 9$ found by our PSD analysis (Section \ref{sec:efd_encoding}).}
    \label{fig:ablation}
    \vspace{-4mm}
\end{figure}

\hTwoSpacePre
\subsection{Evaluation Metrics}
\hTwoSpacePost
\label{sec:metrics}
We compare generated and real datasets using the Compyda framework \cite{necasova_compyda_2024}, which computes morphological features over time from phantom videos and compares the distance between their time series (\textsc{Diff}) also after Dynamic Time Warping (\textsc{DTW}) alignment \cite{paparrizos_survey_2024}. \textsc{Diff} provides an absolute measure of dissimilarity between the two time series over time and evaluates whether the method successfully generates samples within the correct range of values. However, it does not capture whether the features follow the same temporal pattern. The \textsc{DTW} metric measures whether the morphological feature signals exhibits similar dynamics, regardless of frame-by-frame similarity. Since phantom videos can start from different phases of the cells' life cycles, \textsc{DTW} is a more realistic figure of merit.

We select a subset of features computed by Compyda capturing relevant morphological aspects that need to be modelled by a generative process. These features correspond to the phantom's \textit{area} (in pixels), \textit{roundness} (how much it resembles a circle, between 0 and 1), and \textit{elongation} (the ratio between the major and the minor axes). We additionally report \textit{convexity} as the ratio between the phantom's area and the area of its convex hull.

\hTwoSpacePre
\subsection{Comparative Results}
\hTwoSpacePost
We compare our phantom video generation framework against the SSM method used in CellCycleGAN. To ensure a fair comparison, we use the same preprocessing and generative pipeline of \cite{bahr_cellcyclegan_2021}, with the train/test split from Section \ref{sec:datasets}. 
Since this pipeline generates all cell tracking GTs as a single video, we extract each phantom in the same format as the original dataset (Section \ref{sec:datasets}). Additionally, we compare the training set and the test set using the \textsc{Diff} and \textsc{DTW} metrics. This real-against-real comparison represents a reference for the metrics (\textit{Train} in Table \ref{tab:compyda_results}). Since the diffusion process operates in the low-dimensional EFD space ($36 \times 40$ parameters) rather than pixel space ($96 \times 96 \times 40$), our generation is highly efficient. It takes approximately 19 minutes to train Diffusion-TS and 0.09s per phantom video generation (inference plus EFD-to-video conversion time) on a desktop computer equipped with an Intel i9 13900k CPU and a nVidia RTX 3070 GPU. The generation time is in line with the SSM method (0.08s per phantom video on the same hardware setup).  

Our approach achieves higher performance over all metrics computed by Compyda. Table \ref{tab:compyda_results} shows how our synthetic phantom videos exhibit morphological features that are closer to the \textit{Train} baseline with respect to the SSM method. We replicate the generation and evaluation 10 times and report in Table \ref{tab:compyda_results} the mean metric value and the standard deviation of the mean. We additionally perform a t-test between the set of metrics computed over our synthetic dataset and the one generated by the SSM, always obtaining a p-value lower than 1E-10. Notably, in the CTC dataset we achieve an elongation \textsc{Diff} and DTW lower than the real-against-real reference. This indicates that our method can generalize well enough over the morphological properties and avoid overfitting. Fig. \ref{fig:qualitative} shows frames from the test set, from CellCycleGAN's SSM and from our method, highlighting a higher variability of videos generated by our framework. We report frames from the CTC dataset as well, showing that generated phantoms are coherent with the test set evolution in both cases. The contours of phantoms generated by our method are also smoother as a result of the EFD encoding. We also performed an ablation study repeating 10 times generation and dataset evaluation by changing the number of preserved EFD coefficients $\dnyq \in [1, 60]$, showing that few harmonics are enough to capture the cell dynamics, and that the use of a large $\dnyq$ can be detrimental to some features (Fig. \ref{fig:ablation}).
Finally, Fig. \ref{fig:temporal_features} shows that our phantom videos follow closely on the morphological evolution of real data, as the sudden drop in area and convexity, together with the peaks of roundness and elongation, highlight a mitotic event around time step $\TimeStep = 10$. 

\begin{table}[t]
\centering
\caption{We report \textsc{Diff} and \textsc{DTW} over selected morphological features for each model against the test set (lower is better). We report the metrics computed between the real train and the test sets (\textit{Train}), showing that our method generates phantoms that are morphologically close to the reference for both datasets. We report the mean and the standard deviation over experiments replicated 10 times for both our method and the SSM baseline, reporting better metrics in all cases. Rows marked with (*) correspond to a t-test yielding a p-value < 1E-10, showing that the difference between the distributions of our method and SSM is statistically significant.}
\label{tab:compyda_results}
\scriptsize
% \tiny
\setlength{\tabcolsep}{2pt}
\begin{tabular}{|l|r||c|c|c||c|c|}
      \hline
       \multirow{2}{*}{\textbf{Feature}}                 & \multirow{2}{*}{\textbf{Metric}}  & \multicolumn{3}{c||}{CellCycleGAN dataset} & \multicolumn{2}{c|}{CTC dataset} \\  \cline{3-7}
        & & \textbf{\textit{Train}} & \textbf{SSM} & \textbf{Ours}  & \textbf{\textit{Train}} & \textbf{Ours} \\ 
\hline
\multirow{2}{*}{Area}       & \textsc{Diff} $\downarrow$ & $71.17$ & $479.91 \pm 10.66$ & $\mathbf{71.48 \pm  5.26}$* & $47.25$ & $52.40$  \\ 
                    & DTW $\downarrow$  & $32.79 $ & $248.28 \pm  7.17$ & $\mathbf{34.61 \pm  2.91}$* & $23.49$ & $28.45$ \\ \hline %\cline{1-2} \cline{4-7} \cline{9-10} 
\multirow{2}{*}{Roundness}       & \textsc{Diff} $\downarrow$ & $0.04$ & $  0.12 \pm  0.00$ & $\mathbf{ 0.01 \pm  0.00}$* & $ 0.00$ & $ 0.00$  \\ 
                    & DTW $\downarrow$  & $0.01  $ & $  0.09 \pm  0.00$ & $\mathbf{ 0.01 \pm  0.00}$* & $ 0.00$ & $ 0.00$ \\ \hline %\cline{1-2} \cline{4-7} \cline{9-10} 
\multirow{2}{*}{Elongation}       & \textsc{Diff} $\downarrow$ & $0.13$ & $  0.58 \pm  0.02$ & $\mathbf{ 0.15 \pm  0.01}$* & $ 0.18$ & $ 0.11$  \\ 
                    & DTW $\downarrow$ & $0.07  $ & $  0.28 \pm  0.01$ & $\mathbf{ 0.08 \pm  0.00}$* & $ 0.08$ & $ 0.03$ \\ \hline %\cline{1-2} \cline{4-7} \cline{9-10} 
\multirow{2}{*}{Convexity}       & \textsc{Diff} $\downarrow$ & $0.0$ & $  0.06 \pm  0.00$ & $\mathbf{ 0.00 \pm  0.00}$* & $ 0.00$ & $ 0.00$  \\ 
                    & DTW $\downarrow$ & $0.0   $ & $  0.06 \pm  0.00$ & $\mathbf{ 0.00 \pm  0.00}$* & $ 0.00$ & $ 0.00$ \\ \hline %\cline{1-2} \cline{4-7} \cline{9-10} 
\end{tabular}
\end{table}

\hOneSpacePre
\section{Conclusions and Future Work}
\hOneSpacePost
\label{sec:conclusions}
We presented a parametric generative framework for cell phantom video generation using Elliptical Fourier Descriptors and Diffusion-TS. Results demonstrate that our framework effectively learns both morphological properties and temporal dynamics patterns of cells, mimicking the distribution of real data. 
% Currently, we model the morphological evolution of a single track and, if the training set contains mitotic events, we continue the generation of a single daughter's phantom accordingly with the dataset. 
Our work can be integrated into generative pipelines for cell tracking datasets that include composition, movement simulation and texture synthesis. Future work aims at extending our framework to explicitly handle mitotic events (generating both daughter cells during cell divisions) and at generating 3D phantom videos using Spherical Harmonics Decomposition \cite{roa_spherical_2023}.

\bibliographystyle{IEEEbib}
\bibliography{references}

\end{document}